\pdfoutput=1

\documentclass[11pt]{article}

\usepackage{EACL2023}

\usepackage{times}
\usepackage{latexsym}

\usepackage{booktabs}
\usepackage{amsmath}
\usepackage{algorithm}
\usepackage{algpseudocode}
\usepackage{graphicx}
\usepackage{subcaption}
\usepackage{multirow}
\usepackage{svg}
\newcommand{\norm}[1]{\left\lVert#1\right\rVert}
\usepackage[T1]{fontenc}

\usepackage[utf8]{inputenc}

\usepackage{microtype}

\usepackage{inconsolata}

\title{Prototype-Based Interpretability for Legal Citation Prediction}

\author{Chu Fei Luo\thanks{\hspace{2mm}Equal contribution. }\textsuperscript{\hspace{1.2mm}1,2}, Rohan Bhambhoria\footnotemark[1]\hspace{1.2mm}\textsuperscript{1,2}, Samuel Dahan\textsuperscript{2,3}, and Xiaodan Zhu\textsuperscript{1,2}\\
\textsuperscript{1}Department of Electrical and Computer Engineering \& Ingenuity Labs Research Institute \\ Queen's University\\
\textsuperscript{2}Conflict Analytics Lab, Queen's University\\
\textsuperscript{3}Cornell Law School\\
\small{\{\texttt{14cfl,r.bhambhoria,samuel.dahan,xiaodan.zhu}\}\text{\texttt{@queensu.ca}}}} 
\begin{document}
\maketitle
\begin{abstract}

Deep learning has made significant progress in the past decade, and demonstrates potential to solve problems with extensive social impact. In high-stakes decision making areas such as law, experts often require interpretability for automatic systems to be utilized in practical settings. In this work, we attempt to address these requirements applied to the important problem of legal citation prediction (LCP). We design the task with parallels to the thought-process of lawyers, i.e., with reference to both precedents and legislative provisions. After initial experimental results, we refine the target citation predictions with the feedback of legal experts. Additionally, we introduce a prototype architecture to add interpretability, achieving strong performance while adhering to decision parameters used by lawyers. Our study builds on and leverages the state-of-the-art language processing models for law, while addressing vital considerations for high-stakes tasks with practical societal impact.

\end{abstract}

\section{Introduction}
\begin{figure}[t!]
    \centering
    \includegraphics[scale=0.2]{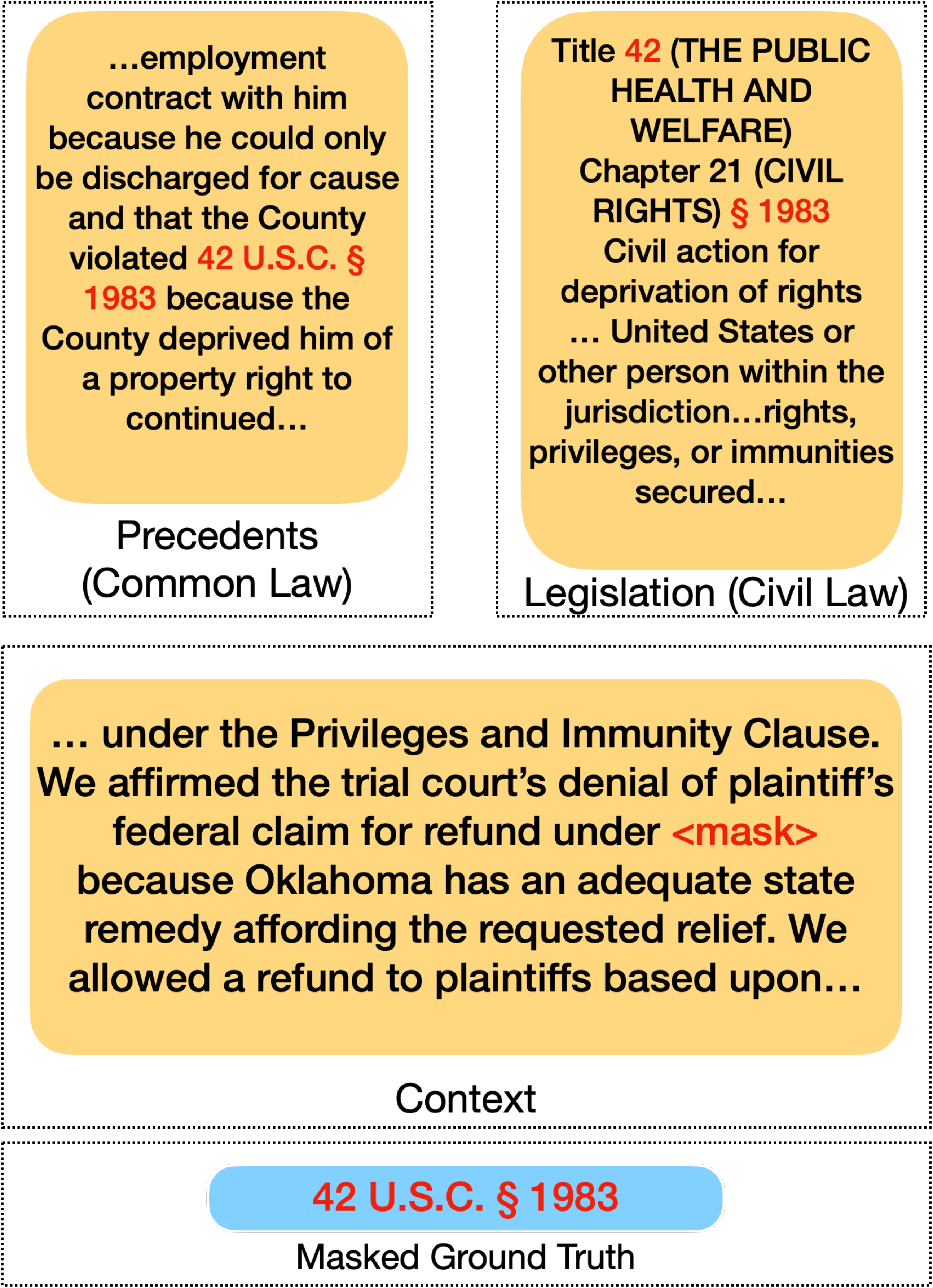}
    \caption{Legal professionals make decisions grounded in (1) Precedents, and (2) Provisions of Legislation. Surrounding context also has a strong influence in determining the ground truth which is represented by <mask>.}
    \label{fig:task}
\end{figure}

Deep learning has made significant progress in the past decade. Researchers have begun applying state-of-the-art methods to problems with extensive social impact, which in turn brings about many critical challenges for these deep learning models. In high-stakes problems, it is essential to understand whether models follow the same reasoning as domain experts or practitioners, and if not, why these models arrive at their final outcomes \cite{arrieta2020explainable}.
Decisions, in other words, often require human comprehensibility for experts to validate and trust their correctness \cite{10.3389/fdata.2021.688969}. 

The field of law is one such high-stakes domain where decision making processes often have a significant societal impact. In this paper, we study a key problem, citation prediction, which aims to predict the most fitting legal citation for a text passage.
Legal citation \cite{paul2022lesicin,xu2020distinguish} is a fundamental component of a lawyer's argument construction process. Unlike in other domains, citations in law are not just used to reference previous works. They serve to indicate ``the nature of the authority upon which a statement is based" \cite{axel1982legal}. 
Court rules go so far as to authorize judges to reject arguments that are not supported by cited authority, and lawyers who appeal on the basis of arguments for which they have cited no authority can be sanctioned \cite{liilegalcitation}. For this reason, any system built to address this high-stakes task must also present evidence to properly support the quality of legal argumentation.



Existing works on legal citation-based tasks \citep{paul2022lesicin} do not adhere to the thought-process employed by lawyers, leading to discrepancies or overlap with existing tasks such as legal judgment prediction. Specifically, existing works primarily predict legal citations solely from the facts of a case. While facts are important to understand a situation, lawyers often develop a legal argument by translating facts in an abstract legal problem first. In other words, citations draw from legal reference to \emph{strengthen the lawyer's position} in court. Many citations are not made to ascertain the nature of the violation. They are instead decided by an understanding of the legal issue the lawyer wants to address, as well as how it relates to prior literature and provisions available in the law. 


In this research, we propose a new definition for this task called \textbf{Legal Citation Prediction (LCP)}. This new approach for citation prediction, illustrated in Figure \ref{fig:task} and described in Section \ref{sec:background}, aims to mimic a lawyer's reasoning by providing prior literature, hereby referred to as \textbf{precedents}, and provisions of legislature, \textbf{provisions}, as input. 
To address the critical issue of interpretability, we extend a prototype-based architecture to better serve the requirements of this task \cite{zhang2021protgnn}. Utilizing prototypes allows the model to formulate representative examples of each citation and draw similarities to legal references. To the best of our knowledge, this work represents the first attempt to use prototype-based interpretability for language model tuning towards a high-stakes legal task.
%

The main contributions of this work include:
\begin{itemize}
    \item Defining a new task of Legal Citation Prediction (LCP), enhanced with feedback from legal experts. We compare performance before and after factoring in legal significance, illustrating the importance and benefits of multidisciplinary collaboration.
    \item Implementing a prototype architecture that approaches this task with the goal of making the model interpretable. It is the first time that this method is being used in a legal application. We introduce a modified loss component which is capable of strengthening parallels to a lawyer's thought-process.
    \item Conducting a thorough analysis on the practicality of this task, providing empirical evidence comparing utility to performance. Additionally, we conduct input perturbations to analyze the learned latent space. As the work is a joint effort of NLP researchers and law practitioners, we also hope it reflects and contributes to learning how machine/deep learning may be more safely deployed in high-stakes fields.
\end{itemize}

\section{Related Work}


\paragraph{Citation Prediction} The task of citation prediction has been extensively studied in academic literature. For scholarly works, citations serve as a method of information search; this is valuable in identifying trends in a given area of research \cite{Yucitationprediction}. The citation frequency of documents can serve as a proxy of that work's influence, which allows for statistical analysis of the trends in the broader community \cite{HOU2019100197}. Citation networks are also used in industry to measure adoption of academic methods \cite{kim2016quantifying}, and in the public sector for allocation of funds by governmental organizations \cite{leydesdorff2019relative}. Most works formulate citation prediction as link prediction on the citation network \cite{Yucitationprediction, liu2019link}, leveraging semantic information from the document texts as well as metadata such as authors and venue \cite{shibata2012link}. 

For legal citation tasks, prior research uses academic citation prediction or information retrieval techniques, which we refer to as Legal Citation Recommendation \cite{huang2021context, dadgostari2021modeling}. \citet{huang2021context}, in particular, explored limiting the context given as input to improve the performance of their works. Although we derive insights from Legal Citation Recommendation research, these studies do not utilize either prior literature or provisions of legislation.

\noindent Another approach is followed by \citet{sadeghian2018automatic}, where the authors construct a heterogeneous citation network from a legal corpus and attempt to predict links as the purpose of a citation. Other works \cite{xu2020distinguish, yang2019legal, wang2019hierarchical} also consider legal citation prediction as a type of Legal Judgment Prediction task, where they identify statutes violation from facts as a proxy to the final judgment of the case. We refer to this formulation of the task as Legal Statute Identification (LSI) based on previous work \cite{paul2022lesicin}. Formally, it is either a multi-label document classification task or an inductive link prediction task that predicts statutes on the basis of the facts of a situation. 
However, while the facts are crucial to formulate a legal argument, citations serve primarily to support a lawyer's arguments. As such, they should not be treated as indicative of the final judgment.
We adopt a multi-label classification setting in this work, but we do not restrict the input to the facts of a case.

\paragraph{Interpretability}

One significant challenge that has been insufficiently addressed in legal citation work is interpretability. Interpretability is a key requirement of AI systems applied to law, as argued in previous studies \cite{gorski-xai-lawyer}. Fulfilling interpretability requirements increases trust in machine learning systems. This, in turn, fosters broader adoption and stimulates further development of applied AI research \cite{rudin2019stop}. In the legal field, there are many problems and tasks that can benefit from NLP models \cite{zhong2020does}. These include legal decision making \cite{bhambhoria2022interpretable}, judgment prediction \cite{Zhong_Wang_Tu_Zhang_Liu_Sun_2020}, and similar charge disambiguation \cite{DBLP:journals/corr/abs-2104-09420}.

\noindent In general, there are many techniques to explain or interpret a model \emph{post-hoc}, i.e. after training \cite{ribeiro2016should, covert2020feature}. However, these methods can be unfaithful to the original model and insufficient for specialized, high-stakes tasks \cite{Luo2022Evaluating, jin2022evaluating}. Consequently, it is important to integrate interpretability into the model's architecture during the training process with \emph{ante-hoc} methods. \citet{bhambhoria2022interpretable}, for example, explored inherent interpretability in a legal decision making task, which enabled lawyers to contribute to model design from an early stage.
Very few works have applied state-of-the-art models to legal citation prediction \cite{paul2022lesicin,xu2020distinguish}, and none have proposed solutions for inherent interpretability. 

\noindent Prototype-based models are successful in various settings, such as few-shot learning, but many works have adapted them to be used for interpretability in both NLP and computer vision \cite{zhang2021protgnn, chen2019looks}. Among existing ante-hoc techniques is ProtGNN \cite{zhang2021protgnn}, a graph-based architecture that makes predictions based on similarity to ``prototypes", i.e. representative training samples for each target label. In this work, we adapt the prototype architecture to provide interpretability for legal citation prediction.
 
\begin{figure*}
    \centering
    \includegraphics[scale=0.28]{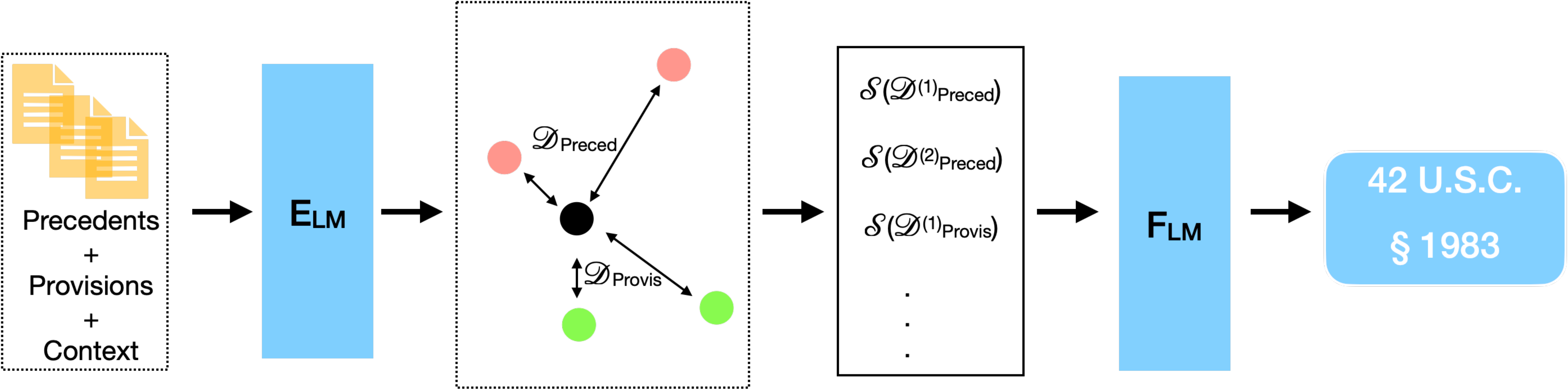}
    \caption{An overview of our architecture. $E_{LM}$ is a RoBERTa-base encoder. The black dot represents the embedding of the input text with surrounding context. Pink dots represent embeddings of precedents. Green dots represent embeddings of provisions. Together, the green and pink dots form the prototypes. $\mathcal{D}$ is used to denote the L2 normalized distance between the prototypes and the input embedding. Calculated scores based on the aforementioned distances is denoted by $\mathcal{S}(\mathcal{D}^{i}_{Preced/Provis})$. The scores are passed through a feedforward layer, $F_{LM}$ to obtain the top scoring provision.}
    \label{fig:architecture}
\end{figure*}
\section{Prototype-Based Legal Citation Prediction}
We propose a prototype-based architecture to address the task of Legal Citation Prediction. The full task is described in Section \ref{sec:background}, and our architecture is shown in Figure \ref{fig:architecture}. Prototype-based architectures add interpretability by basing their predictions on similarity to ``prototypes", which are representative training samples for each target label. Our proposed architecture enhances interpretability and grounds the model decision in a lawyer's thought process via a customized loss function that considers \textbf{precedents} and \textbf{provisions} as comparison points. For our LCP task, we use a combination of automatically discovered prototypes for precedents and manually chosen prototypes for provisions. We use vanilla fine-tuning as a baseline, where we append a linear classification head to a pre-trained language model and fine-tune all parameters with multi-class cross-entropy loss, only using the case text as input. At this stage, we also explore performance tradeoffs of task configurations such as limiting text spans and the number of target labels. We also incorporate expert feedback from initial results to adjust the experiment parameters. Then, we fine-tune from the base pre-trained model while adding our prototype-based custom loss objective, encouraging an interpretable latent space organized around precedents and provisions. We explore performance trade-offs and the learned embedding space through perturbations. 

\subsection{Defining Legal Citation Prediction}
\label{sec:background}
We define Legal Citation Prediction (LCP) as a multi-label classification based on the input text, as well as the provisions and precedents for our target citations. For a passage of text $x \in X$, and a set of possible target citation labels $L$, where $|L| = n$, the goal is to predict the subset of appropriate target citations $y \in L$, represented as an $n$-dimensional vector.

A critical motivation for LCP is the thought process lawyers undertake when finding citations for their work. Similar to scholarly citations, lawyers attempt to find the most relevant precedent to strengthen their own arguments \cite{savelka2021discovering}. When lawyers present a legal argument and judges write opinions, they package their interpretation of what the law is and how it should be applied to a given situation. During this process, lawyers make reference to statutes, regulations, court rules, as well as prior appellate decisions they believe to be pertinent and supporting. We define \textbf{provisions} as pieces of written law, such as statutes, regulations, and court rules, and \textbf{precedents} to be prior appellate decisions made in court. Various legal systems apply provisions and precedents in varying scales of importance. 
For example, common law is distinguished by its use of prominent appellate decisions, also known as caselaw.
These two components have distinct characteristics, in the same way the definition of a word and its appearance in a sentence can inform the word's usage in different ways. Existing works have strictly used prior appellate decisions, or strictly used the source text, but we theorize making both available can provide more context to citation prediction. In addition to the input text $x$, we also make available the target provisions to be cited, $Provis$, and relevant precedents, $Preced$, taken to be other text passages with the same target citation in the training set. We automatically sample representative precedents in this work, described in Section \ref{sec:training}.

\begin{algorithm}[t]
\caption{Prototype formulation algorithm for prototypes $p_j \in P$, and target citation labels $l_j \in L$. $E_{LM}$ denotes a language model encoder, $\cos$ is cosine similarity, and $\operatorname{Cluster}$ is any clustering algorithm that can produce centroids $C$.}\label{alg:cap}
\begin{algorithmic}[1]
\Require Training dataset $\{x_i, y_i\}^n_{i = 1} \in X$, Possible cited provisions $l \in \operatorname{L}$
\State Encode all samples $f(x_i) = E_{LM}(x_i)[0]$
\For{$l \in \operatorname{L}$}
    \State $X_l = x_i \in X\ |\ y_i(l) = 1$
    \State $C_l = \operatorname{Cluster}(f(X_l))$
    \For{$c_{l,j} \in C_l$}
        \State $x_{l,j} = \operatorname{argmin} _{x \in X_l} \cos(f(x), c_{l,j})$
        \If{$\cos(f(x_{l,j}), c_{l,j}) > s_{min}$}
            \State $p_j = f(x_{l,j})$
        \Else
            \State $p_j = c_{l,j}$
        \EndIf
    \EndFor
\EndFor
\end{algorithmic}
\end{algorithm}

\subsection{Training Prototypes} 
\label{sec:training}

The prototype construction process is described in Algorithm \ref{alg:cap}. First, we encode the entire training set into the latent space. Then, for each target provision in the legislation, we find the subset of training samples that contain a citation to that text, denoted by $X_l$. We then cluster $X_l$ with k-means, taking the centroids as prototype candidates $C_l$. For each candidate, we try to locate the closest training sample by cosine similarity $x_{l,j}$. If the similarity exceeds a chosen threshold $s_{min}$, we update the candidate to match the embedding of $x_{l,j}$. Otherwise, we take the candidate as the final prototype to be used in training.

For a batch of \(n\) samples, with input embeddings \(f(x)\) for input \(x\), and a feed-forward classification head \(c\), the loss is denoted by $\mathcal{L}$ in Equation \ref{eq:lmloss}. The input embedding \(f(x)\) is taken as the CLS token, and similarity scores $\mathcal{S}$ are calculated with the similarity score described in Equation \ref{eq:distance}. We add L2 normalization to the distance metric from previous work \cite{zhang2021protgnn}, where $p_k$ is the prototype, $h$ is the embedding output of $f(x_i)$, and $\epsilon$ is a regularizing weight. Also, we use the standard binary cross-entropy loss for multi-label classification.
\begin{equation}
\label{eq:lmloss}
\begin{split}
    \mathcal{L} = \frac{1}{n} \sum_{i=1}^{n} \operatorname{BCELoss}\left(c \circ \mathcal{S} \circ f\left(x_{i}\right), y_{i} \right)\\
    + \lambda \mathcal{D}_{preced} + \delta \mathcal{D}_{provis}
\end{split}
\end{equation}
\begin{equation}
\label{eq:distance}
\begin{split}
\mathcal{S} = \operatorname{sim}\left(p_{k}, h\right)=\left| \log \biggl(\norm{\frac{\left\|p_{k}-h\right\|_{2}^{2}+1}{\left\|p_{k}-h\right\|_{2}^{2}+\epsilon}}^2\biggr)\right|
\end{split}
\end{equation}

All precedents are represented in the loss as $\mathcal{D}_{Preced}$ with a similar formulation to the previous work, \cite{zhang2021protgnn}. In this work, we extend the loss by adding a new term, $\mathcal{D}_{Provis}$ to represent legislation provisions. We also establish the existing loss terms with coefficients of $\lambda$ to represent precedents, as shown in Equation \ref{eq:preced}.

\begin{equation}
\label{eq:preced}
\begin{split}
    &\mathcal{D}_{preced} = \lambda_{1} \frac{1}{n} \sum_{i=1}^{n} \min _{j: p_{j} \in P_{y_{i}}}\left\|f\left(x_{i}\right)-p_{j}\right\|_{2}^{2} \\
    &+\lambda_{2}(-\frac{1}{n} \sum_{i=1}^{n} \min _{j: p_{j} \notin P_{y_{i}}}\left\|f\left(x_{i}\right)-p_{j}\right\|_{2}^{2}) \\
    &+\lambda_{3}\sum_{k=1}^{C} \sum_{\substack{i \neq j \\ p_{i}, p_{j} \in P_{k}}} \max \left(0, \cos \left(p_{i}, p_{j}\right)-s_{\max }\right)\\
\end{split}
\end{equation}
\begin{equation}
\label{eq:def}
\begin{split}
    \mathcal{D}_{provis} = \frac{1}{n} \sum_{i=1}^{n} \min _{j: d_{j} \in D_{y_{i}}}\left\|f\left(x_{i}\right)-d_{j}\right\|_{2}^{2}
\end{split}
\end{equation}

\noindent The $\lambda_{1}$ term is used to encourage embeddings to move closer to a prototype cluster of their class, where $f(x_i)$ is the embedding representation of the input $x_i$, $p_j \in P_{y_i}$ is the set of prototypes belonging to all provisions of legislation cited $y_i$, and $n$ is the batch size. In contrast, the $\lambda_{2}$ encourages embeddings to move away from prototypes of different classes, scored similarly to $\lambda_{1}$ but using $p_j \notin P_{y_i}$. The $\lambda_{3}$ term moves prototypes of the same class further away from each other, punishing prototypes that have a cosine similarity above a threshold $s_{\max}$. $\mathcal{D}_{Provis}$ is defined similarly to the $\lambda_{1}$ term, as shown in Equation \ref{eq:def}, and serves a similar purpose of encouraging input embeddings to be closer to the provision source text embedding.  


\section{Experimental Settings}

\begin{figure}[t!]
    \centering
    \includegraphics[width=0.7\linewidth]{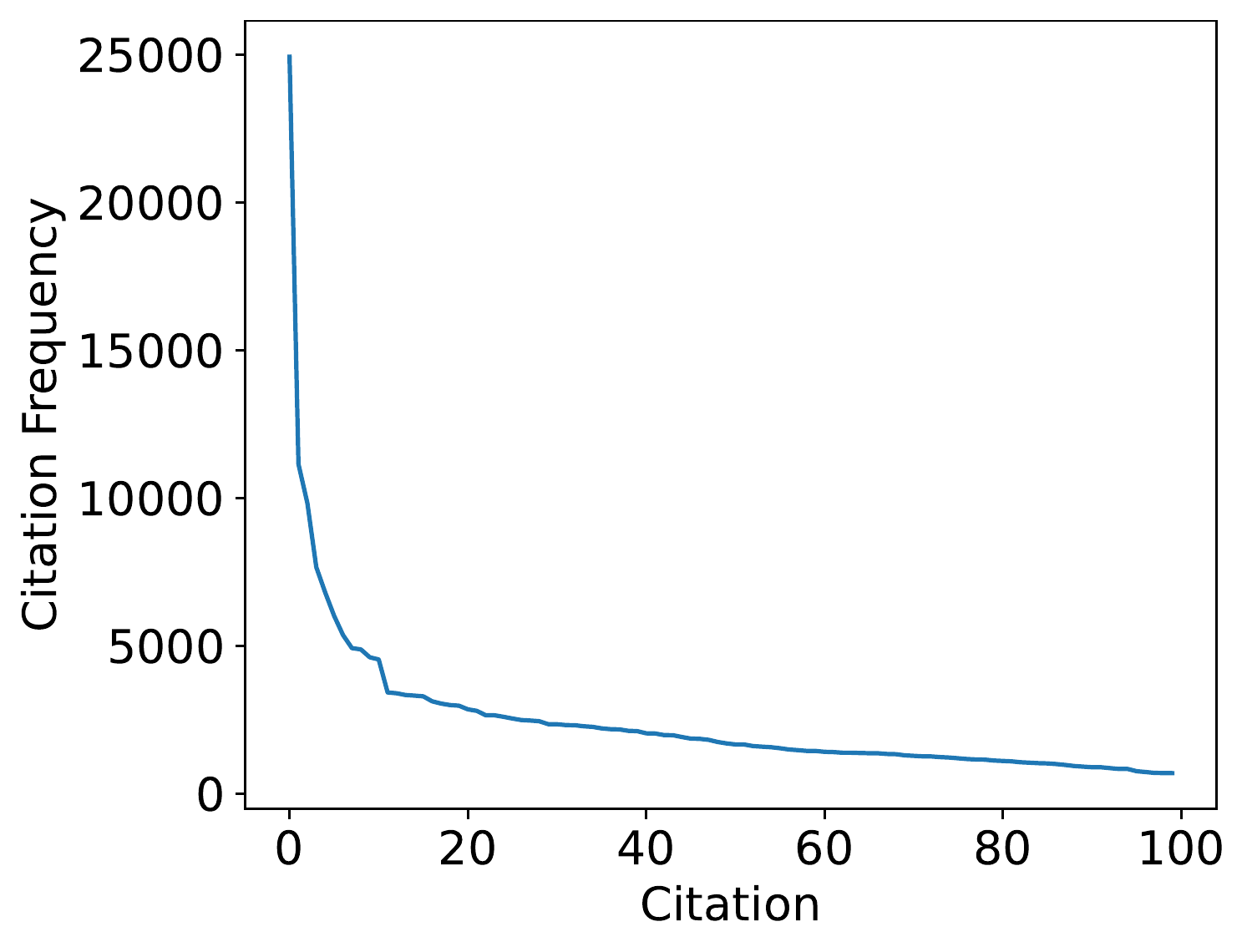}
    \caption{Distribution of documents citing the top 100 most frequently occurring provisions.}
    \label{fig:dataset}
\end{figure}

\begin{table}[t]

\renewcommand\arraystretch{0.9}

    \centering
    \setlength{\tabcolsep}{3pt}

    \resizebox{0.75\linewidth}{!}{

     \begin{tabular}{p{2cm}|ccc}
  \toprule
\textbf{U.S. Code} & \multicolumn{1}{l|}{Ann. 1} & \multicolumn{1}{l|}{Ann. 2} & Ann. 3 \\ 
    \midrule
     \textbf{42 \S 1988}   & \multicolumn{1}{c|}{1.5}& \multicolumn{1}{c|}{1} & 2 \\
      \textbf{28 \S 1404}    & \multicolumn{1}{c|}{1} &\multicolumn{1}{c|}{1.5} & 2.5 \\
    \bottomrule
  \end{tabular}

    }

  \caption{Expert feedback study results. Three legal experts (Ann. 1, Ann. 2, Ann. 3) were provided 4 samples citing a provision of U.S. code, and asked to rate the relevance of the passage text to the citation on a scale of 3.}
  \label{tab:ratings}

    \vspace{-1em}

\end{table}
\subsection{Encoder Model Selection} 
\label{sec:model}
Preliminary experiments are summarized in Table \ref{tab:prelim}. Please refer to Appendix \ref{sec:prelim} for model details. We use \emph{vanilla fine-tuning} for all preliminary experiments. With just the case text as input, we append a linear classification head to a pre-trained language model (depicted as $E_{LM}$ in Figure \ref{fig:architecture}), and fine-tune all parameters with multi-class cross-entropy loss. At this stage, we also explore performance tradeoffs of task configurations such as limiting text spans and the number of target labels. We choose LegalBERT for our architecture as it outperforms RoBERTa and has equivalent performance to Longformer for lower computational cost.
\begin{table*}[t]

\renewcommand\arraystretch{0.9}

    \centering
    \setlength{\tabcolsep}{2pt}

    \resizebox{0.7\linewidth}{!}{

    \begin{tabular}{p{2.cm}|cc|cc|cc}

    \toprule

\multirow{2}{*}{\textbf{Model}}                                                  & \multicolumn{2}{c|}{5 labels}                                           & \multicolumn{2}{c|}{20 labels}    & \multicolumn{2}{c}{100 labels}           \\ 

        \cmidrule{2-7}

         &Macro-f1 & Micro-f1  &Macro-f1 & Micro-f1 &Macro-f1 & Micro-f1  \\

        \midrule

        RoBERTa & 51.6 & 49.5 & 55.0 & 59.6 & \textbf{34.4} & \textbf{45.2}  \\
        LegalBERT & \textbf{54.4} & 49.4 & 55.1  & \textbf{60.7} & 32.9 & 44.8 \\
        Longformer & 53.6 & \textbf{51.3} & \textbf{56.0} & 58.8 & 33.7 & 44.9  \\                

      \bottomrule


    \end{tabular}

    }

  \caption{A summary of preliminary experiments to choose the encoder pretrained language model and sets of target citations. We report Macro- and Micro-F1 scores, and all experiments report a single run.}
  \label{tab:prelim}

\end{table*}
\subsection{Dataset}

We built a dataset of court opinions, hereby referred to as the PACER dataset, constructed from United States federal court documents curated by the Free Law Project\footnote{\url{https://free.law/}}. This data has been used in previous works \cite{dadgostari2021modeling}. However, we download and preprocess the data from scratch. These documents are derived from 1276 jurisdictions in the United States of America, ranging from the federal Supreme Court to local district or municipal courts. We downloaded the files via the Free Law Project's CourtListener bulk API on February 10, 2022. For this work, we focus on predicting provisions of the U.S. Code. The provision source text associated with each target citation is retrieved from the Legal Information Institute (LII) maintained by Cornell Law School\footnote{\url{https://www.law.cornell.edu/}}. The gold labels are automatically extracted from the text via regex. We remove documents that do not contain any U.S. Code citations, and filter for documents that contain at least one of the top 100 most frequently cited subsections, for a total of 175,741 documents. Each opinion cites an average of 3.02 U.S. Code provisions, and each provision has an average of 5308.51 citing opinions. This is divided into a 80:5:15 train, validation, and test split ratio. The dataset label distribution is long-tailed, with the most frequently cited U.S. code appearing more than twice as often as the next. The label imbalance can be observed in Figure \ref{fig:dataset}.

\paragraph{Expert Feedback}
\label{sec:legal}
We sought to obtain feedback on the PACER dataset and prototype discovery from legal experts. After conducting preliminary experiments, further described in Section \ref{sec:model}, we performed one iteration of our prototype discovery process on the best-performing checkpoints. We encoded the train set with RoBERTa \cite{liu2019roberta}, clustered the embeddings, then found example cases in the training set closest to the cluster centroids by cosine similarity. 

We then showed four examples and their provision source text to three legal experts, including the source text of the provision and the full court opinion, and asked them to rate the relevance of the provision citation to the original case. The legal experts were asked to rate each example on a scale of 3, where 3 is highly relevant and 1 is completely irrelevant, and the results are summarized in Table \ref{tab:ratings}. Overall, the average rating of the four examples was 1.5 --- i.e. the relevance of the prototypes is relatively low. Responses for one example are shown in Table \ref{tab:survey} of Appendix \ref{sec:feedback}. 

The legal experts stated in a follow-up interview that this was due to the target citations $L$. We chose frequency of citation as an indication of importance, but citations serve different purposes, as mentioned in previous work \cite{sadeghian2018automatic}. Other legal citation tasks manually choose prediction targets based on significance \cite{paul2022lesicin}, or use citations that serve a specific purpose, like caselaw \cite{dadgostari2021modeling}. Many of the citations we automatically targeted were procedural; they define proper procedures in court proceedings, such as appropriate legal fees, but are not relevant to the legal argument being made. The lawyers gave low ratings because the citations were not valuable prediction targets.

With the assistance of a legal expert, we manually removed procedural citations from the top 100 automatically chosen targets. A second expert was consulted to validate the filtering and sort ambiguous categories. We kept definitions as relevant, since they do not form a legal basis but are important to building an argument. We considered government regulations, such as allowance administration expense, to be adjacent to procedural citations and removed them as well. Of the top 100 citations, 55 (55\%) of them are procedural; of the top 20, 15 (75\%) are procedural.

\subsection{Surrounding Text Span Context}
\label{sec:context}

\begin{table*}[t]

\renewcommand\arraystretch{0.9}

    \centering
    \setlength{\tabcolsep}{2pt}

    \resizebox{0.7\linewidth}{!}{

    \begin{tabular}{p{2.cm}|cc|cc|cc}

    \toprule

\multirow{2}{*}{\textbf{Context}}                                                  & \multicolumn{2}{c|}{20 labels}                                           & \multicolumn{2}{c|}{100 labels}    & \multicolumn{2}{c}{45 labels}           \\ 

        \cmidrule{2-7}

         &Macro-f1 & Micro-f1  &Macro-f1 & Micro-f1 &Macro-f1 & Micro-f1  \\

        \midrule

        N/A & 55.1 & 60.7 & 32.9 & 44.8 & 43.4 & 50.9  \\
        $\pm$4 & 66.7 & 68.9 & 50.3  & 58.8 & 68.7 & 73.7 \\
        $\pm$2 & \textbf{68.9} & \textbf{71.1} & \textbf{55.7} & \textbf{62.0} & \textbf{69.5} & \textbf{74.9}  \\                

      \bottomrule


    \end{tabular}

    }

  \caption{A summary of vanilla fine-tuning performance with different input contexts and target citation sets. N/A refers to the setting where no context is taken, and we only classify on the first 512 tokens of a document.}
  \label{tab:baselines}

\end{table*}


The definition of the Legal Citation Prediction task implicitly suggests we are only concerned with the portions of a document that are relevant to the citation. This differs from focusing on the course of events that led to a potential violation of a law. Also, the 512 token limit in some of our pre-trained language models like RoBERTa is a significant limitation when parsing longer legal documents. We theorize that the surrounding context within a document is more important for a citation, which may contain opinions and arguments alongside facts of the situation, as discussed in previous works \cite{Yucitationprediction}. To address this issue in our work, we filter document sentences by the \emph{surrounding context} of our target citations, i.e. taking \(n\) sentences before and after a citation sentence. For example, $\pm$2 implies for a sentence of the input that contains a citation $s_c \in x_i$, we retain the 2 sentences before and after in the sequence $\{..., s1, s2, s_c, s3, s4,...\}$, resulting in 5 sentences total. Table \ref{tab:context} in Appendix \ref{sec:dataset} illustrates how this preprocessing significantly reduces document length. With $\pm$2 context, the mean document length is below the token limit. We do not always classify all 100 target citations. Some experiments remove labels to handle dataset imbalance, and others to reflect the feedback of legal experts, as described in Section \ref{sec:legal}. In the case where a document does not contain any of the target citations, we randomly sample sentences until there is a minimum of 15 selected.

\begin{table*}[t]

\renewcommand\arraystretch{0.9}

    \centering
    \setlength{\tabcolsep}{7pt}

    \resizebox{0.8\linewidth}{!}{

    \begin{tabular}{p{3.6cm}|cc|cc}

    \toprule

            \multirow{2}{*}{\textbf{Experiment Setting}}                                                  & \multicolumn{2}{c|}{20 labels}     & \multicolumn{2}{c}{45 labels}                           \\ 

        \cmidrule{2-5}

         &Macro-f1 & Micro-f1  &Macro-f1 & Micro-f1 \\ 
         
         \midrule
$Preced$ & 69.0 \small{(+0.1)} & 72.9 \small{(+1.8)} & 69.3 \small{(-0.2)} & 74.4 \small{(-0.5)}\\
$Preced + Provis$ & 73.2 \small{(+4.3)} & 73.4 \small{(+2.3)} & 65.9 \small{(-3.6)}  & 71.4 \small{(-3.5)}  \\ 
 \midrule
Keyword Masking & 42.5 \small{(-30.7)} & 52.4 \small{(-21.0)} & 57.7 \small{(-8.2)} & 63.6 \small{(-7.8)} \\
Random Masking & 73.7 \small{(+0.5)} & 74.8 \small{(+1.4)} & 68.7 \small{(+2.8)} & 72.2 \small{(+0.8)} \\
Freezing Encoder & 66.9 \small{(-6.3)} & 69.2 \small{(-4.2)} & 64.0 \small{(-1.9)} & 72.4 \small{(-1.0)}    \\

      \bottomrule


    \end{tabular}

        }

  \caption{Model ablations and perturbations with our prototype-based architecture. $Preced$ refers to training with only $\mathcal{D}_{Preced}$ loss from Eqn. \ref{eq:preced}, and $Preced + Provis$ refers to training with both $\mathcal{D}_{Preced}$ and $\mathcal{D}_{Provis}$ loss terms. The numbers in brackets for the two model ablations indicate the deviation from vanilla fine-tuning. All perturbations are performed training with $Preced + Provis.$, and the numbers in brackets indicate deviation from the $Preced + Provis$ prototype results}.
  \label{tab:ablations}

\end{table*}

\section{Results and Analyses}


\paragraph{Baseline} Results of our baseline experiments with vanilla fine-tuning are shown in Table \ref{tab:baselines}. The purpose of these experiments is to determine the optimal number of surrounding sentences to provide as input context during training, and also to examine the effects of target labels on performance. Removing rarer classes helps alleviate the challenge of the long-tail imbalance in the dataset, as discovered in previous works \cite{ma2020learning}. However, one of the challenges of law is the vast citation network; therefore, it is important to investigate model performance over as many possible citations as possible. From the preliminary experiments in Table \ref{tab:prelim}, we observe that all models demonstrate slightly higher prediction accuracy with 20 labels compared to 5, but the performance decreases significantly with 100 labels.To this end, we continue further experiments with \textbf{20 labels} and \textbf{100 labels}. We add a \textbf{45-label} setting with only non-procedural citations based on expert feedback in Section \ref{sec:legal}.

Under the 20-label setting, we observe that inputs without context filtering and naively taking the beginning of a document results in the worst performance. Providing context of 4 sentences preceding and following the required citation, denoted by $\pm 4$, leads to better results. Finally, by including $\pm 2$ sentences for context, we obtain the best performance. As expected, we observe lower F1 scores for 100 labels due to the increase in rare classes. For subsequent experiments, we maintain 20 labels as the baseline for the following reasons: 1) We observe strong empirical evidence for this setting, and 2) The number of labels corresponds to a reasonable number of outcomes which can correspond to prior filtering by legal professionals, and our system would help with disambiguation.

\noindent The performance of 45 labels without context follows a similar trend, and performance falls between 20 labels and 100 labels. However, once we provide surrounding context to the model, we see a significant increase in performance, with $\pm 4$ and $\pm 2$ both improving over the baseline by 23-25 points in both Macro- and Micro-F1. Compared to the other settings that exhibit 15-22 points in F1 improvement, it is clear that context is more important with legally relevant citations, or procedural citations are more likely to be mentioned in the first 512 tokens.
\begin{figure*}[t!]%
  \centering
  \begin{subfigure}[c]{0.47\linewidth}
    \centering
    \includegraphics[trim={0.5cm 0.5cm 0 0},clip,width=\linewidth, height=5cm]{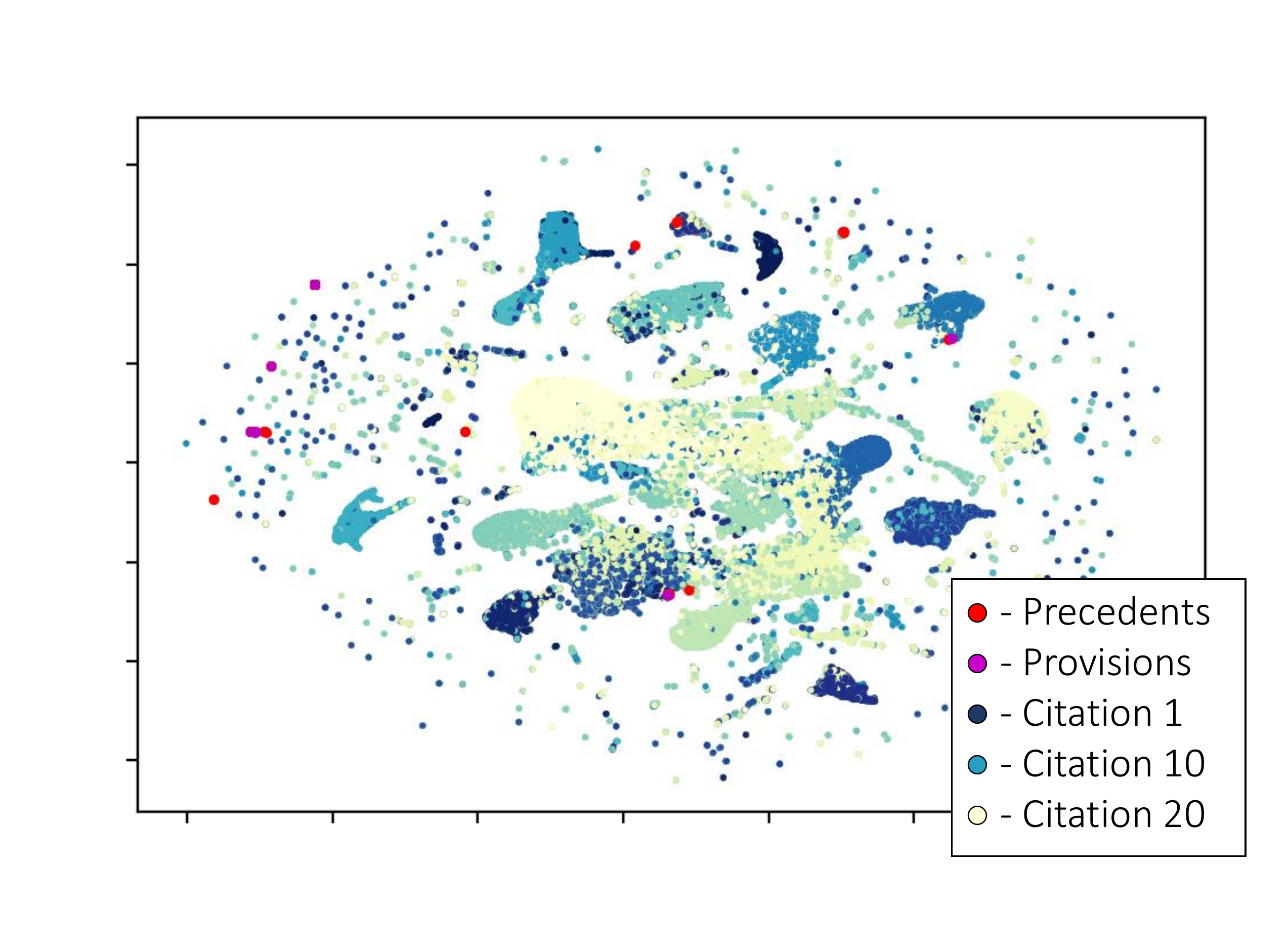}
    \caption{Projection after vanilla fine-tuning.}
    \label{fig:plotft}
  \end{subfigure}
  \begin{subfigure}[c]{0.47\linewidth}
    \centering
    \includegraphics[trim={0 0.5cm 0 0},clip,width=\linewidth, height=5cm]{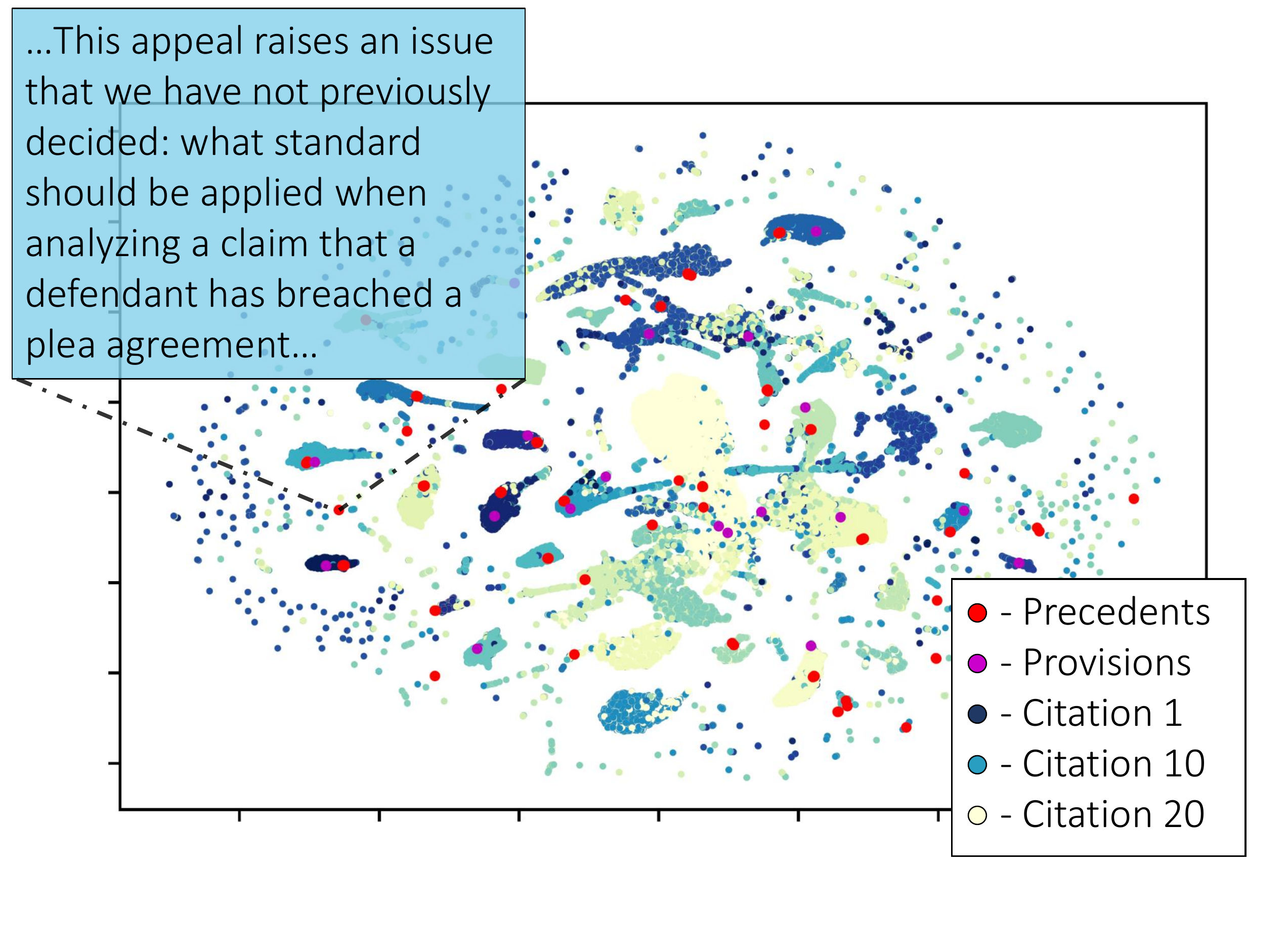}
    \caption{Projection after fine-tuning with $\mathcal{D}_{Preced}$ + $\mathcal{D}_{Provis}$.}
  \end{subfigure}
  \caption{Latent space projections after our prototype-based fine-tuning, colour-coded by citation label. Red dots are precedent-sourced prototypes, magenta dots are provision-based prototypes, and every other colour represents samples from a different target citation.}
  \label{fig:plots}
\end{figure*}
\paragraph{Prototype-based Model} Next, we test the prototype-based model based on the best performing baseline configuration of $\pm 2$ labels, summarized in Table \ref{tab:ablations}, reporting results on 20 and 45 labels to compare performance with and without considerations to legal significance. We perform a simple model ablation: first, we train from the base model with only the $\mathcal{D}_{Preced}$ loss term. Next, we add the $\mathcal{D}_{Provis}$ term that incorporates provision source text. 

Over these experiments, the prototype-based loss results in comparable performance to vanilla fine-tuning while only tuning on the $\mathcal{D}_{Preced}$ term. However, the 20-label task (75\% procedural citations) sees dramatic improvements from the $\mathcal{D}_{Provis}$ term, outperforming vanilla fine-tuning by 4 points in Macro-F1. Conversely, the $\mathcal{D}_{Provis}$ term with the same hyperparameters seems to hinder the performance of the 45-label task, decreasing by 4 points in Macro-F1.


\paragraph{Perturbations} To further validate the feature importance of the provisions, we perform several perturbations from our $Preced + Provis$ model. We attempt two settings: 1) \textbf{Keyword Masking}, i.e. replacing the input keywords with the [MASK] token, and 2) \textbf{Random Masking}, where we randomly mask 15\% of the input tokens. For keyword masking, we use a statistical keyword extractor, YAKE \cite{yake}, to extract 20 ngrams, up to $n=2$, from the provision text.

\noindent All results are summarized in Table \ref{tab:ablations}. \textbf{Keyword Masking} reduces the model performance significantly compared to random masking for both the 20-label and 45-label setting, but the effects are stronger with 20 labels. This implies that citations with less relevance to the legal argument have more similarity to their source provision, and our distance-based classification sees increased performance. It also explains why legal experts found little value in predicting these citations; if it is easy to predict a citation from keywords, the task would be trivial to a lawyer. Conversely, citations with more significance to a legal argument seem to have more abstract relationships to the surrounding context and to other citing documents, and constraining them to the provision text has an adverse effect on the performance. This also explains why random masking offers better performance, as this step reduces spurious noise \cite{wang-etal-2022-identifying}.

\paragraph{Vanilla Fine-tuning vs. Prototypes} Additionally, we compare vanilla fine-tuning to prototype-based training by freezing the latent space, as denoted by the \texttt{Freezing Encoder} experiment in Table \ref{tab:ablations}. We perform prototype discovery, encode the provision text, then train our classification head $c$ without updating the prototype embeddings or language model parameters. It is interesting to note the performance decrease ranges from 1.0 to 6.3. In other words, holding everything else constant, re-organizing the latent space with our prototype-based loss results in a 1.0-6.3 point increase in F1 score.

We visually inspect the learned information by projecting the latent space of the 20-label models. We reduce the dimensionality of the embeddings with UMAP \cite{mcinnes2018umap} to produce 2D projections shown in Figure \ref{fig:plots}. Comparing the baseline to our prototype-based loss, the latter is visually more well-organized; there are fewer outliers at the edges of the latent space and clear clusters of cases corresponding to different citations. 

After further investigation into the architecture, we observed there was \emph{only one prototype} discovered for each target citation after vanilla fine-tuning. The remaining cluster centroids did not meet the cosine similarity threshold. This likely increases redundancies in the activations, which makes it easier to find patterns in embedding distances. However, the prototypes are arbitrarily located in the latent space, so the model might be learning spurious prototype activation patterns.

\section{Conclusion}

We study Legal Citation Prediction (LCP), a problem that serves as a foundational function for decision making in modern legal systems with societal impact. In our work, we built an inherently interpretable prototype architecture that is compatible with any language model encoder, and explains its decisions based on similarity to precedents and provision text of the target citation. We automatically discover prototypes from the training set for each target citation, which reduces the need for expert input, and make predictions based on an input's similarity to these prototypes. Through empirical study, we show strong evidence of our architecture towards LCP, offering more interpretability than vanilla fine-tuning for equivalent performance.

We also demonstrate that leveraging a combination of precedents and the target provisions' texts in the during model training results in comparable or greater performance to the baseline language model, although the effectiveness of our full architecture depends on the legal nature of the target citations. Compared to vanilla fine-tuning, our model's latent space visibly separates different citation embeddings into distinct groups. It is possible to apply our distance-based classification head to an encoder trained with vanilla fine-tuning, but the classification is likely based on spurious features. In practice, this system could be extended to any citation target that has prior examples and source text, such as prior cases (i.e. caselaw). Interpretability is a crucial requirement for deploying transparent AI systems, and we encourage more work in this direction for applications with significant social impact.

\section*{Limitations}

We observe two main legal limitations for this project. First, it has limited practical use for non-lawyers seeking legal help, also called self-represented litigants. In fact, any system providing legal citations, both precedent or statutory provision, to an untrained lawyer will be of very little use, and even harmful. It is hard to imagine in what context this might be used by non-lawyers considering that they might not be able to translate facts into a legal problem. That being said, many direct-to-public legal applications have emerged recently, and many of these applications do provide insightful legal information along with the legal sources \cite{lawsociety,dahan2020case}. While these applications have raised concerns as to their legality, notably with the issue of unauthorized practice of law, many regulators including in Canada, the United States and Europe have cautiously supported the development of AI-power technology for the general public. 

Second, several lawyers (especially appellate) have surprisingly expressed concerns regarding ``the Googlization of legal databases" \cite{vaidhyanathan2011googlization}. While they recognize the advantages of intuitive AI non-Boolean research, they claim that these algorithms are not superior when it comes to locating a more obscure appellate case law, to help win a case. It has even been argued that Boolean logic remains faster and more efficient because it does not lead to missed case. According to this view, while the ``Googlized" legal database may quickly locate important caselaw especially if decided by a higher court, it can miss less obvious cases \cite{susan-aall}. In our work, this challenge translates to the long-tail problem for legal citations, and our use of embedding distance encourages matching based on semantic similarity. In other words, we only look for the most obvious citations, which correlates to higher performance on easier (procedural) citations and lower performance on harder (non-procedural) citations. Our work does not sufficiently address this problem, so we encourage more proficient information retrieval or prototype discovery methods in the future.

From a deep learning perspective, the main limitation is due to the use of k-means clustering in our implementation of the system. There were several points of instability noted during the training process, which we theorize is largely due to the initializations of the k-means clustering algorithm. When the prototypes are initialized, the corresponding terms in the loss function have a strong influence on the cross-entropy loss, which leads to model collapse. Even when the prototypes are initialized properly, the loss function overfits to the prototypes after several updates but does not provide improvement in the classification performance, which is why we choose the best model by validation macro F1 instead of validation loss.




\section*{Ethics Statement}
\paragraph{Intended Use}
We see at least two applications for legal practice. First, this system could serve as predictive text drafting application for legal memo and judicial opinions. This application would recommend a list of citations - both precedents, statutes and even secondary literature - that is the most relevant to the legal problems and concept discussed in the memo or opinion. Second, such application may also be integrated into legal databases, such as Westlaw or LexisNexis. While these databases have been working on new algorithms based on non-Boolean keyword searches with more intuitive AI features, more work is needed when it comes to finding the most relevant citation. 

\paragraph{Failure Mode}
Although the task of citation prediction is high-stakes and key in a lawyer's decision making process, risks associated with system failure are mitigated due to the system's enhanced interpretability. Since this system is intended for lawyers, and also classifies based on similarity to previous works, a user would be able to leverage their expertise to validate the decision. If the chosen provision text does not match the legal argument the user had in mind, they can easily examine the similarity to other available citations, or discard the system's decision entirely.

\paragraph{Misuse Potential}
As mentioned in the paper, there is a high potential for people to confuse legal citations with legal judgment, and people can leverage the discovered citations to directly decide a ruling. This system could be misused in that way similar to previous models used in the industry, such as COMPAS \cite{Kirkpatrick2017ItsNT}.

\section*{Acknowledgements}
The research is in part supported by the NSERC Discovery Grants, New Frontiers in Research Fund, and the Research Opportunity Seed Fund (ROSF) of Ingenuity Labs Research Institute at Queen's University. We also acknowledge feedback from David Liang and Solinne Jung from the Conflict Analytics Lab, as well as the ACL review committee, that helped us improve the work.

\bibliography{anthology,custom}
\bibliographystyle{acl_natbib}

\appendix

\section{Additional Implementation Details}
\label{sec:implementation}

\subsection{Training Parameters}
All experiments were run on 11GB Nvidia 2080 GPUs. We used an initial learning rate of 2e-5, weight decay of 0.01, batch size of 8, and trained the system for 20 epochs. The training pipeline and models were implemented using Huggingface and Pytorch python libraries, and all pre-trained language model checkpoints were also downloaded from Huggingface's online repository. The total runtime is approximately 30 hours for 20 epochs using the BERT-base variants, but we observe convergence in the loss by 10 epochs. Additionally, we select the best model by validation macro-F1 score with our prototype model instead of validation loss. All experiments are reported as a single run.

Clustering for prototype discovery was implemented with the PyKeops (Python Kernel Operations) library \footnote{\url{https://www.kernel-operations.io/}}. We used k-means clustering on the embedding space with k=5, clustered on cosine distance, and we re-cluster the prototypes every 5 epochs. We tested k=3 and k=5 for clustering, and chose k=5 as the best performing. This result aligns with the findings of previous work \cite{zhang2021protgnn}. We tested other configurations, such as euclidean distance instead of cosine, but these settings gave the best performance. For the $\mathcal{D}_{Preced}$ weights, we use the same $\lambda$ values as described in \cite{zhang2021protgnn}, where $\lambda_1 = 0.10$, $\lambda_2 = 0.0005$, and $\lambda_3 = 0.001$. Then, we chose $\delta=0.10$ for $\mathcal{D}_{Provis}$, and a minimum cosine similarity $s_{min}$ of -1.

\subsection{Preliminary language modelling results}
\label{sec:prelim}
We experimented with three models:
\begin{itemize}
    \item \textbf{RoBERTa-base} \cite{liu2019roberta} (110M parameters) --- An optimized version of BERT, which is an encoder-only pre-trained language model.
    \item \textbf{LegalBERT} \cite{chalkidis-etal-2020-legal} (110M parameters) --- A variant of BERT domain adapted to law by pre-training on court cases.
    \item \textbf{Longformer} \cite{Beltagy2020Longformer} (149M parameters) --- An optimized transformer architecture with sparse attention for a longer context window (4096 tokens vs. 512 in BERT).
\end{itemize}
\begin{table}[!t]
  \centering
  \begin{tabular}{lll}
  \toprule
     \textbf{U.S. Code}   & \textbf{\# citations} & \textbf{\# documents}\\
    \midrule
    42 \S\ 1983     & 64524 & 31246 \\
    11 \S\ 523 (a)      & 22377 & 7676 \\
    28 \S\ 1331     & 18419 & 13967 \\
    28 \S\ 157 (b)      & 16517 & 12319 \\
    42 \S\ 1981     & 13190 & 6178 \\
    \bottomrule
  \end{tabular}
  \caption{Statistics on frequency of citations for the top 5 most frequently cited U.S. codes. U.S. Code citations are formatted [Title] \S [Section] [(optional) Subsection]. Each citation can appear in a document multiple times, but even counting documents alone, there is a significant imbalance between the different labels.}
  \label{tab:dataset}
\end{table}

\begin{table}[!t]
  \centering
  \begin{tabular}{llll}
  \toprule
     \textbf{Context}   & \textbf{Min.} & \textbf{Max.}    & \textbf{Mean} \\
    \midrule
    N/A & 5 & 120593 & 1973.83 \\
    $\pm$4\ & 3 & 39848 & 825.96 \\
    $\pm$2\ & 3 & 21005 & 466.06 \\
    \bottomrule
  \end{tabular}
  \caption{RoBERTa token count statistics with varying context spans.}
  \label{tab:context}
\end{table}


\subsection{Preprocessing} 
During the training process, we use regex expressions to remove all HTML web formatting, links, U.S. Code citations, and Supreme Court citations. Non-ASCII characters are also removed, but everything else is preserved as they did not cause a noticeable decrease in performance. Additionally, when predicting fewer than 100 citations in later experiments, we again remove any documents that do not contain the target citations. For our train set of 141237 documents, this step results in 50567 documents for experiments on the top 5 citations, 90243 for top 20 citations. We also retrieve surrounding text spans in different configurations as described in Section \ref{sec:context}. 

\section{Additional dataset analysis}
\label{sec:dataset}
\begin{figure*}[t!]%
  \centering
  \begin{subfigure}[c]{0.36\linewidth}
    \centering
    \includegraphics[width=\linewidth]{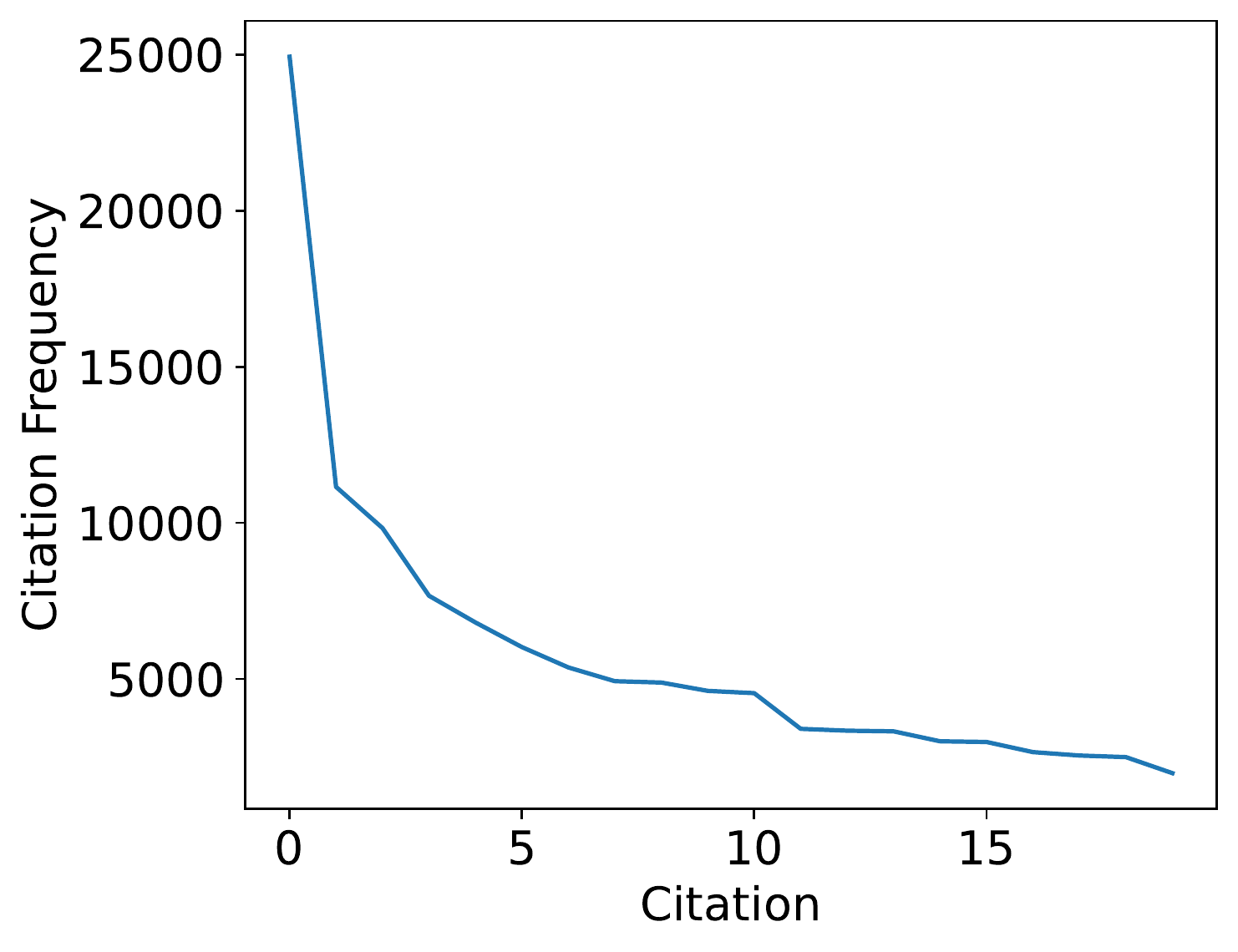}
    \caption{Frequency of citations for the 20-label setting.}
  \end{subfigure}
  \begin{subfigure}[c]{0.36\linewidth}
    \centering
    \includegraphics[width=\linewidth]{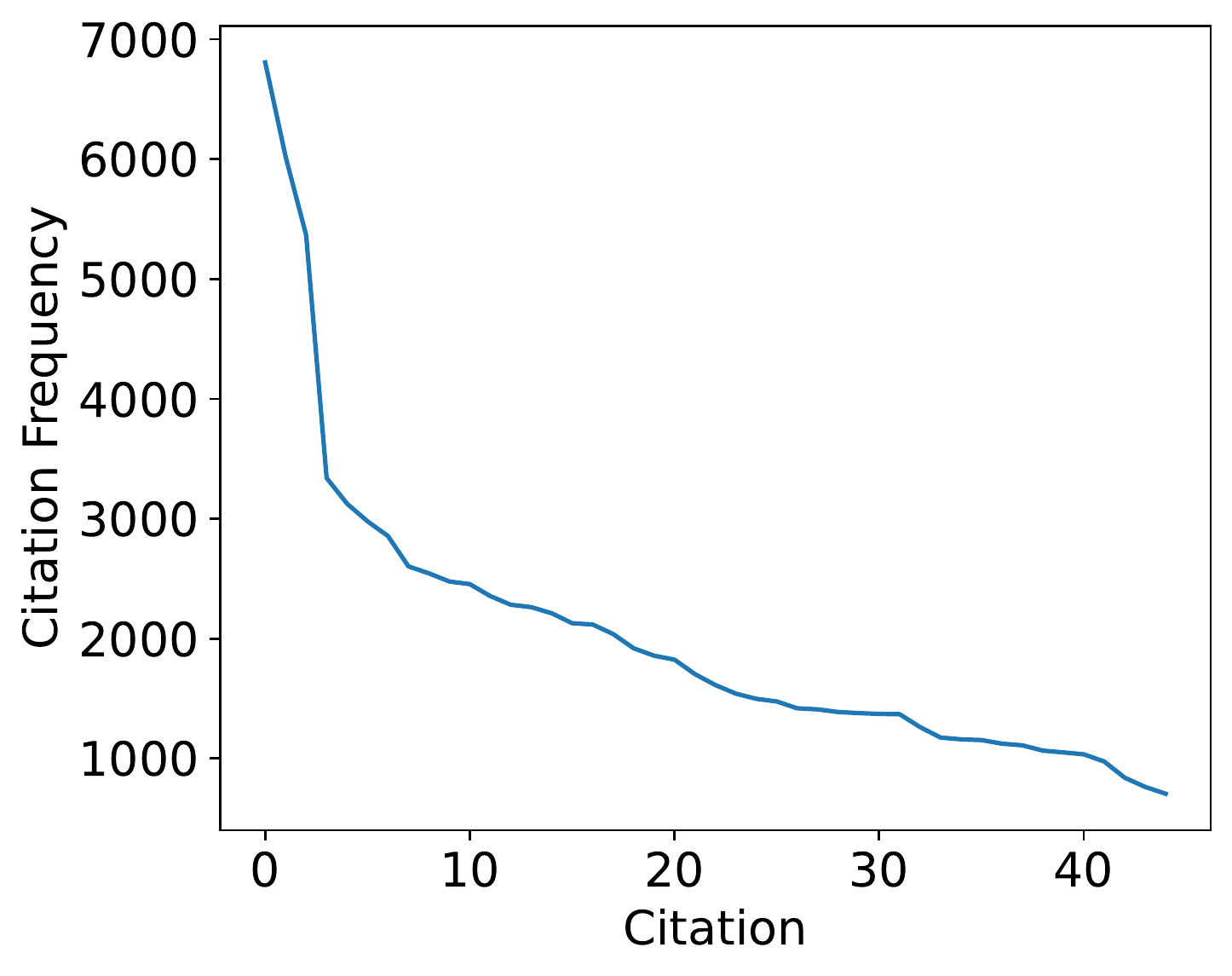}
    \caption{Frequency of citations for the 45-label setting.}
  \end{subfigure}
  \caption{Additional plots of citation frequency, demonstrating that every tested citation set exhibits a long-tail distribution.}
  \label{fig:counts}
\end{figure*}

The long-tail nature of the dataset stays consistent across the different target label settings we reported, as shown in Figure \ref{fig:counts}.

\section{Expert Feedback}
\label{sec:feedback}
We recruited three volunteer legal experts through our institution for our brief feedback study. They have approximately 20 years of combined experience and were able to converse freely to share opinions. While we did not provide compensation for this specific study, they are active collaborators and receive salary or course credits for this interview.
\begin{table*}[!t]
    \centering
    \begin{tabular}{|l|l|}
    \hline
    \textbf{Citation}       & 42 U.S. Code § 1988                                                               \\ \hline
    \textbf{Provision} & Proceedings in vindication of civil rights \\
    
& (a)Applicability of statutory and common law...\\

&\textbf{(b)Attorney’s fees}... the court, in its discretion, may allow the prevailing party, other \\
& than the United States, a reasonable attorney’s fee as part of the costs, except that in \\
& any action brought against a judicial officer for an act or omission taken in such officer’s\\
&judicial capacity such officer shall not be held liable for any costs, including attorney’s \\
&fees, unless such action was clearly in excess of such officer’s jurisdiction. \\

& (c)Expert fees...                                         \\ \hline
    \textbf{Precedent}    & <s>MEMORANDUM-DECISION AND ORDER KAHN, District Judge. Presently \\
    & pending is a motion by Plaintiff Association of International Automobile \\
    &Manufacturers, Inc. ("AIAM") for \textbf{attorneys fees} pursuant to Fed.R.Civ.P. 54(d) (A) \\
    & and <mask>. Plaintiff asserts that it is the prevailing party in an action brought under \\
    & and is thus presumptively entitled to such fees. Ass'n v. Cahill ("AAMA II"), <mask>, \\
    & reprinted at 1997 U.S.C.C.A.N. 1077, 1388. The remaining two prongs are also clearly \\
    & established. First, the right created by 209(a)'s prohibition is demonstrably not so \\
    & vague and amorphous that its enforcement will strain judicial competence...\\ \hline
    \textbf{Expert 1}    & 2: It seems somewhat relevant but hard to say as I have no expertise in this area. \\
    &Also, I wonder whether the model is a Legal-Bert and saw the whole case.                 \\ \hline
    \textbf{Expert 2}    & 3: The relevant part is the part about the attorney fees   \\ \hline
    \textbf{Expert 3}    & 1: The provision deals with compensating a party for attorney fees in order to enforce \\
    & an action. The sample, however, is dealing with a party/state's right to enforce \\
    &emission standards. \\ \hline
    \end{tabular}
    \caption{An example of a legal citation, the contents of the provision, a court opinion discovered by clustering the latent space, and comments from the 3 legal experts. 2 of the 3 legal experts mention the few sentences regarding attorney fees as relevant, but the third brought up the issue of these citations being ultimately separate to the case's contents.} 
    \label{tab:survey}
\end{table*}
\end{document}